\documentclass{article}
\usepackage{graphicx} 
\usepackage{url}
\usepackage{listings}
\usepackage{xcolor}

\usepackage{tcolorbox}

\usepackage{subcaption}

\usepackage[utf8]{inputenc}
\usepackage{booktabs}
\usepackage{longtable}
\usepackage{hyperref}
\usepackage[margin=1in]{geometry}
\usepackage{enumitem}
\usepackage{tikz}
\usetikzlibrary{decorations.pathreplacing,positioning,arrows.meta}

\hypersetup{
    colorlinks=true,
    linkcolor=blue,
    urlcolor=blue,
    citecolor=blue
}
\lstdefinestyle{yaml}{
    basicstyle=\ttfamily\small,
    breaklines=true,
    frame=single,
    backgroundcolor=\color{gray!10},
    keywordstyle=\color{blue},
    stringstyle=\color{red},
    commentstyle=\color{green!60!black},
    showstringspaces=false,
    xleftmargin=2em,
    numbersep=8pt
}

\title{Granite.Trust Policy Tools: \\Shareable, Actionable Policies for Generative AI Applications}
\author{Nathalie Baracaldo, Nicolas Mello, Kush R. Varshney, Heiko Ludwig, \\ Kate Soule, David Cox \\ 
\small IBM Research}

\begin{document}

\maketitle

\begin{quote}
\textit{Your application, your policies, your model: Using these Actionable Policy Tools is a first step to define policy your way}     
\end{quote}

\begin{abstract}
When it comes to safety policies for generative AI, one size does not fit all. Each organization and use case needs to mitigate different risks depending on the application context, regulatory environment, organizational values, and user personas. Yet, existing policy specification approaches are designed for traditional access control and fail to capture the nuances of GenAI application: the enforcement of content-based constraints.

We present two contributions to address this gap:
\textbf{(1)} the \textit{Actionable Policy schema}, a YAML-based format for specifying what model responses can and cannot contain.
The schema enables \textit{exception-based policy governance},
proposing exceptions to track policy violations;
\textbf{(2)} \textit{synthetic data generation pipeline} that produces policy-aligned training data for model alignment and testing,
and \textit{a set of tools} to help define the schema and enforce policy. Together, these enable organizations to specify policies once and enforce them throughout the GenAI application lifecycle: from model alignment to runtime monitoring. The Actionable Policy schema, example policies, and tools are available as open source:
\begin{center}
\url{https://github.com/ibm-granite/granite.trust.policy-tools}
\end{center}
We welcome new ideas, contributions and feedback.

\end{abstract}


\section{Introduction}
Generative AI (GenAI) applications pose novel safety and other risks beyond traditional enterprise applications caused by their generative nature where responses and actions are generated by large language models (LLMs) trained on a large corpus of data. 
At the same time, GenAI applications in enterprises are scrutinized for their behavior by regulators, customers, the general public, and various stakeholders within an enterprise with the objective of meeting safety, security, regulatory or business expectations.
These expectations are frequently defined as \textit{policies}.

Policy compliance is required of actions taken in regulated environments. In the GenAI era, this fundamental requirement has not changed, yet the nature of policy has. Defining a risk posture and resulting policy is needed to materialize a set of procedures that ensure a compliant GenAI application development and deployment lifecycle. Real-world enterprise and consumer GenAI applications and agents require their own specific policies.

\begin{figure}[htbp]
\centering
\includegraphics[width=0.9\columnwidth]{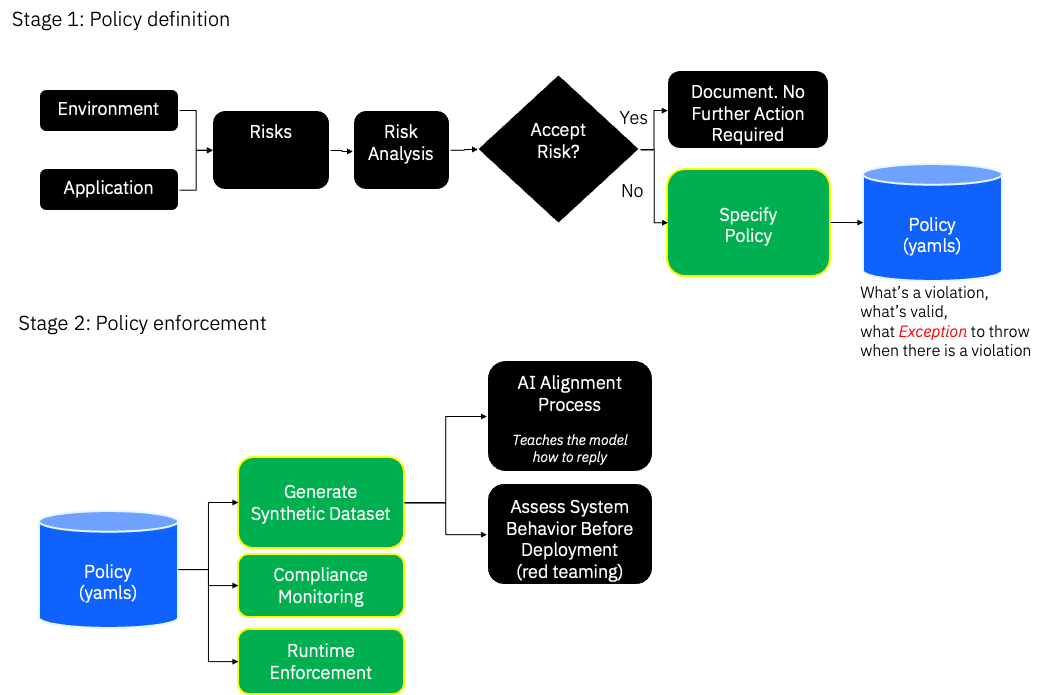}
\caption{Overview of the proposed policy framework. During Stage 1, the policy is defined as a result of a risk management process to mitigate relevant risks. Risks may change, and so do policies. The resulting policy is used in Stage 2 for a variety of applications including generating synthetic datasets for model alignment and assessing system compliance before deployment, compliance monitoring and runtime enforcement. 
}
\label{fig:overview}
\end{figure}

\begin{figure}[htbp]
\centering
\includegraphics[width=0.8\columnwidth]{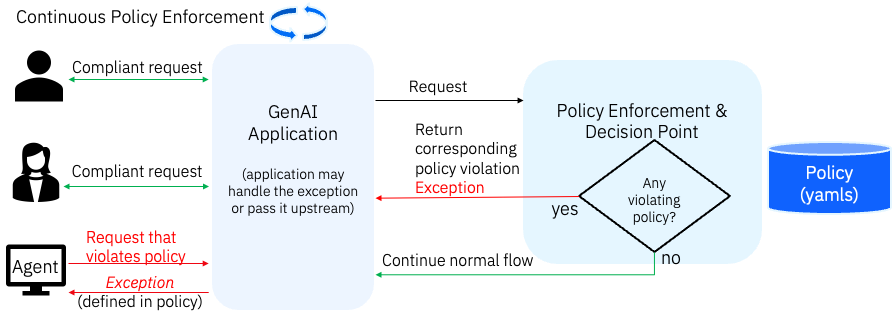}
\caption{The policy format provides a concrete definition of \texttt{Exception} types that allow tracking  policy violations and handle them properly either in the application or upstream (in another agent). The third example in the figure shows the exception is passed to the agent so the agent can handle it appropriately.}
\label{fig:policy_exception}
\end{figure}

While risk management in GenAI applications is a topic of taxonomization (e.g., NIST AI RMF, MIT AI Risk Repository, IBM Risk Atlas),
there is not yet a unified way to specify \textit{policies} that can influence the behavior of a GenAI application.
While some research has studied how to adequately reply to a diverse set of questions \cite{ashktorab2025s},
how to respond \textit{in adherence to policy} when a risk materializes remains a largely unsolved challenge.

Ideally, a suitable policy specification should enable the design of a very specific set of controls and oversight. Figure \ref{fig:overview} illustrates how risk and policy relate to each other (Stage 1) and how defining a policy can directly impact a set of procedures during the design and development of models through model alignment, testing and red teaming (Stage 2). Detailed requirements are presented in Section \ref{sec:requirements}. As seen in Stage 1 of the figure, risks arise from the interaction between the GenAI application and the environment where it will be deployed. Risk analysis methodologies aim to uncover those risks and manage them in a way that will lead to a successful application. During this process, risks may be \textit{accepted} in which case simply documenting that they were uncovered is enough. However, for unacceptable risks, a mitigation policy is required.

\subsection{The Policy Gap in GenAI}
The gap between risk identification and actionable policy is what the proposed Granite.trust Policy Tools address.
They provide a way to specify policy for GenAI applications that has not been covered before.
Traditional access control policies (XACML, OPA/Rego, Cedar) answer questions like: \textit{Can user X access resource Y?} These are binary decisions based on attributes. GenAI policies must additionally answer fundamentally different questions:

\begin{itemize}
    \item \textit{What content can the model include in its response?}
    \item \textit{How should the model decline a harmful request?}
    \item \textit{What exceptions should be logged when a policy boundary is crossed?}
    \item \textit{How can we communicate policy violations?}
\end{itemize}

In addition to the challenges above, we find that each organization has a distinct set of policies that are not shared with others.
\textit{One size doesn't fit all.}
For that reason, having a format that allows stakeholders, such as legal experts and application owners, to collaborate with AI experts to express such policies is a must.
Only by defining a clear policy is it possible to ensure that a GenAI application can be properly overseen.

Policy can also improve oversight of aspects that are not covered by default set up in models and guardrails.
Relying on pre-existing guardrails and model alignment does not solve the problem. 
Guardrails are tailored to principles that are not easy to verify,
and as a user, there is little flexibility on what can be identified as risk.
Fine-tuning models, an alternative to identify policy violations, requires acquiring datasets that closely reflect the policy of the organization. However, the likelihood of finding these datasets online is not high. For example, there is a limited number of datasets to address requests that involve competitors. As a result, having a way to create datasets that mirror the desired policy behavior would improve the policy compliance landscape. The proposed policy tools can help tailor those controls to very specific use cases.

\subsection{Contributions}

This paper makes the following contributions:

\begin{enumerate}
    \item \textbf{Policy Schema} (Section~\ref{sec:unit-format}): A YAML-based format for specifying content-based policies that is human-readable yet machine-enforceable. The schema captures what responses can and cannot contain, how to decline requests, and what exceptions to raise.

    \item \textbf{Policy-Driven Synthetic Data Generation} (Section~\ref{sec:synthetic-data}): A pipeline that generates adversarial prompts and safe responses aligned with a given policy, enabling model fine-tuning and compliance testing. This pipeline ensures that no matter what your policy may be, you can generate data for model alignment, generating guardrails or red team your application.  

    \item \textbf{Exception-based policy governance}: Our policy definition provides a way to track policy violations across application boundaries (Figure \ref{fig:policy_exception}). Our follow up paper will cover this aspect in detail. 

    \item \textbf{Complementary Tools:}
    We evaluated the format suitability for policy specification, we identify and implemented additional tools to improve the process of defining policies. 
\end{enumerate}

Section~\ref{sec:requirements} presents the requirements that drove our design. Section~\ref{sec:unit-format} presents the proposed schema and in Section~\ref{sec:synthetic-data} a pipeline to generate synthetic data that complies with the policy requirements. 
In Section~\ref{sec:evaluation}, we present the results of some usage evaluation by legal and technical users. Section~\ref{sec:tools} highlights some of the tools we have developed to facilitate policy definition and compliance. We conclude the report in Section \ref{sec:conclusions}. 

We release additional accountability tools for the lifecycle of a GenAI application at \url{https://github.com/ibm-granite/granite.trust.policy-tools} under an Apache 2.0 license.

\section{Requirements and Related Work} \label{sec:requirements}

\begin{quote}
    \textit{What you can't measure, you can’t control. A clear policy definition is necessary for effective evaluation and governance}
\end{quote}

We created a Policy format to specify policies that can directly influence the behavior of large language models and GenAI applications. These are the requirements: 

\begin{enumerate}
    \item \textbf{Human Understandable:} 
    Risk assessment and management requires collaboration between technical teams, legal counsel, and compliance officers. The policy format must be readable by non-programmers. A lawyer should be able to specify how the application should behave for a particular risk without learning complex policy languages.

    In our experience, we found that it is easy to fall into discussions that lead to policies that are too abstract to be actionable. The proposed schema should ensure that discussions among stakeholders lead to \textit{actionable} enforceable and verifiable outcomes of what can and cannot be replied.
    
    \item \textbf{Machine Actionable:} While human readability is essential, the policy must also be precise enough to drive automated tools. Free-form natural language policies lead to ambiguous interpretations that cannot be reliably enforced or tested. The specification must be concrete enough to enable:
    \begin{itemize}
        \item Synthetic data generation for model alignment
        \item Automated compliance testing
        \item Runtime policy enforcement
    \end{itemize}

    \item \textbf{Versioned for Compliance:} 
    Compliance frameworks (e.g., ISO 42001, EU AI Act) require tracking policy changes over time. The format must support versioning to enable audit trails and compliance certification. During IBM Research's ISO 42001 certification audit for its Granite model development process, this versioning was required. In addition, it has helped keep track of what datasets we had generated with what definitions and how new risks have been mitigated. 
        
    \item \textbf{Shareable Across Organizations and Agents:} 
    Organizations should be able to share policies with partners, regulators, and the community. A standard format enables policy comparison, conflict detection, and collaborative policy development.

    \item \textbf{Exception-Aware:} When a policy boundary is crossed, the system must know how to respond. Inspired by software exception handling, policies should define typed exceptions that propagate through the application stack, enabling consistent violation handling.

\end{enumerate}

\subsection{Why Existing Approaches Fall Short}

There are multiple techniques that aim to minimize risks.

\subsubsection{Risk Management Taxonomies and Frameworks}
To this date, a variety of risk taxonomies have been proposed. These taxonomies play an ever-important role in identifying potential risks that need to be considered before deploying a GenAI application. They are the first stepping stone to reason about safety and security. Notable risk frameworks include AI Risk Atlas \cite{bagehorn2025ai}, OWASP top 10 \cite{owasp-llm}, the MIT repository \cite{MIT:risk:repo:2026}, the AIR taxonomy \cite{zeng2024ai} and \cite{eisenberg2025unified:credoAI,slattery2024ai}.
Additional taxonomies have been proposed by benchmarks such as AILuminate \cite{mlCommons:ailuminate}, HELM \cite{helm-safety}, AirBench \cite{zeng2024airbench}, BBQ \cite{parrish2022bbq}, among others. 

Yet, risk management cannot end there. 
A next step is to assess each risk to define if measures to mitigate it are needed and what are those measures. In some cases, a risk may require deploying a mitigation strategy, while in others, accepting the risk is enough.
This requires defining how to address risks through a well-defined process to determine if they should be accepted or addressed in a particular fashion.
Example risk management frameworks for cybersecurity include the OCTAVE \cite{alberts2003octave}.
In the AI space, Credo AI presented steps to generate mitigation techniques for AI mitigation \cite{eisenberg2025unified:credoAI}. Their approach allows for general risk mitigation policies such as use access controls. However, the proposed \textit{policy packs} are closed source making it difficult to fully compare. In contrast, our approach is open source and is much more specific to the point where we can very clearly specify how to generate \textit{policy}-driven datasets that can serve to align or verify the system.
Another approach is the OSCAL Compass \cite{oscal-compass}, a cloud-native compliance space that allows the specification of policies in a language that enables users to understand policies. Our approach targets a different use case: \textit{how to reply to users requests in policy and verify compliance}.
Our policy schema was inspired by LlavaGuard \cite{helff2024llavaguard} which was designed for vision models and included around eight policies. We augmented the policy specification including additional items such as versioning to enable ISO 42001 certification \cite{iso42001}, a high-level description of the risk, the definition of an \textit{Exception} designed for easy GenAI application reliable error handling and agentic interaction, and a desired remediation. 

Risk mitigation techniques also include red teaming approaches where LLM models and GenAI applications are verified to ensure flaws are detected. 
These approaches, e.g., \cite{ares,munoz2024pyritframeworksecurityrisk,petri,crescendo:russinovich2025great}, usually generate synthetic prompts tailored to specific LLM vulnerabilities for example role playing.
They frequently require \textit{seeds} which are sample prompts of offending content, or \textit{scenario} definitions that describe what to test for.
Red teaming approaches can be combined with the proposed policy to tailor evaluation (only flag or prioritize issues based on whether they violate policies) and find target problems in the areas where the policies are specified. 

\subsubsection{Relying on the Model Default Security and Safety}
A first option is relying on the \textit{pre-baked} security and policy offered by models and guardrails.
LLMs are typically aligned for security and safety reasons. 
Guardrails are additional models that aim to regulate the model outputs.
These are typically good first lines of defense; however, 
these approaches have been trained to comply with general safety principles, constitutions or model specification the model providers have defined \cite{granite,padhi2024graniteguardian,openAi:modelspec:2025,gemma4,shieldgemma2,purplellama,bai2022constitutionalAI,anthropic:new:constitution:2026}.
Typically, these descriptions are vague (lack transparency) and organizations and end-consumers do not have control of the policies or principles used for alignment.
Over-refusal \cite{rottger2024xstest,panda2024llm,ashktorab2025s} is another potential pitfall related to relying on pre-defined security and safety. Applications should lead the way in which some sensitive questions are answered using policy rather than generic language to satisfy use cases.
Additionally, relying on pre-aligned models is not enough in cases where fine tuning is applied. This is known as the 
\textit{fine-tuning breaks alignment} challenge \cite{qi2023finetuning:breaks:alignment} and requires additional protections beyond standard alignment \cite{wang2025invariance:unlearning}. 

Recently, the Granite Guardian model was enhanced to be promptable for specific risks brought by the user without requiring fine-tuning  \cite{guardian4:1}. Our proposed approach is complementary: it is possible to use that model capability to feed Granite Guardian a policy for enforcement.\footnote{Granite Guardian and scripts to evaluate conversations can be found in the Policy Tools repository.}

\subsubsection{Fine Tuning, Steering and Prompt-based Mitigation Methods}
Having a way to change how an LLM  answers based on specific use cases is extremely relevant and dependent on the particular use case.
For example, answers provided in a touristic GenAI application should be tailored differently than those for an HR chatbot or an application for kids.
Methods in this category include
modifying the model's weights, or generating LoRA \cite{hu2022lora} or aLoRA \cite{greenewald_activated_2025} adapters. Optimization methods such as RLHF \cite{christiano2017rlhf}, SFT \cite{sft-llms-blog,ding2025improvedSFT}, DPO\cite{rafailov2023dpo}, GRPO \cite{shao2024grpo}, unlearning \cite{liu2024rethinkinguUlearning,wang2025invariance:unlearning,unlearning:book:2026,wang2025rethinking}, constitutional AI \cite{bai2022constitutionalAI}, multi-human-value alignment palette (MAP) \cite{wang2024map}, in context learning \cite{brown2020languageICL}, steering \cite{garcia2025refusal} and many others have been proposed.
All these approaches manipulate the model towards a desired state and require datasets that exhibit the desired and undesired behavior.

\textbf{Open-source dataset:}
One alternative is to rely on open source datasets, e.g., \cite{bai2022training:anthropic:rlhf,ghosh-etal-2025-aegis2,brahman-kumar2024coconot}. These datasets usually include samples to target \textit{risks} that are deemed relevant.
However, we found that in practice, some risks are not covered. For example, it is not easy, if possible at all to find a readily available dataset that aligns to the desired policies to discuss competitors in a GenAI application.
In some other cases, the quality of the data may not be as good or may be limited\footnote{As an example, dataset \cite{ghosh-etal-2025-aegis2} contains samples for child safety, however, the number of such samples is in the single digits, which is not enough to train a model}.
Finally, it is not clear what policy guidelines were used to generate replies to publicly available datasets.
For these reasons, having a way to generate datasets according to the desired properties is extremely important.

\textbf{Human-generated datasets:}
We also observed that some approaches are human-labor intensive. 
RLHF-based approaches for alignment require users to spend time verifying how a particular model is working and giving ``live" feedback for each reply \cite{christiano2017rlhf,gemma-model-alginment}.
These approaches require a human in the loop and only consider \textit{general feedback}, for example \textit{be more creative}.
Our policy-driven data generation can be integrated with these approaches by consolidating feedback in the policy description itself and generating verification samples.
Datasets for unlearning can also be generated with this policy description.

\textbf{Synthetic Data Generation}
Creating human datasets is quite expensive. Therefore, 
synthetic data generation has received a lot of attention \cite{xu2024magpie,li2023camel,dgt,wang2024codeclm,helff2024llavaguard,ding2023enhancing,yin2023dynosaur,li2024synthetic,synthetic:document:anthropic:2025,lambert2024tulu,achintalwar2024alignment}.
These approaches were designed for generating datasets in fields such as math, physics, instruction following and others.
However, they were not designed to adhere to policies.
For example, MAGPIE \cite{xu2024magpie} generates samples by querying a larger LLM that is assumed to already follow properly the right alignment. While this is adequate for certain fields, it is not suitable for policy verification.
Dromedary \cite{sun2023dromedaryCox} provides an approach to guide synthetic data generation by principles without policy verification. The approach proposed in this paper allows to specify very specific behaviors.
In \cite{padhi2024value}, policy documents are used as seeds to generate synthetic data for model alignment. However, unlike our approach, their policies are not directly verifiable by relevant stakeholders.
The closest approach to ours is LlavaGuard \cite{helff2024llavaguard}. As we mentioned before, they propose generating synthetic data restricting what it can contain. However, their policy format is uniquely designed for synthetic data generation, while the policy proposed in this paper goes beyond data generation: it also specifies how to address policy violations that may occur during the application's runtime. 

\subsubsection{Handling Safety Exceptions at Runtime}
With an extremely fast pace with which new models are published, and the amount of agents using a diverse set of models (sometimes in an opaque way), we need a way to verify at runtime policy in a way that we can \textit{i)} specify the compliance requirements once, and \textit{ii)} verify for all models. 
The policy description can serve as a verification tool.
Unlike evaluations such as \cite{evaluation:llama:2026},
the policy specifies a much clearer and detailed set of requirements. Finally, we notice a lack of approaches to make policy failures a first-class citizen.
For real applications, having a way to specify and treat policy violations in a cohesive way to the best of our knowledge has not been addressed by existing approaches.


\section{Policy Schema} \label{sec:unit-format}

In this section, we present the Policy schema (version 1.0).
We collaborated with members of IBM's internal governance function during the design and evaluation of the schema. The policy tools effort was informed by the process of developing actual policies.
  
\subsection{Design Principles}

The schema embodies four design principles:
\begin{enumerate}
    \item \textbf{Evolving Risk Landscape:} Risks and regulations change over time, especially in a nascent technology such as GenAI. The schema should enable keeping track of policy changes.
    
    \item \textbf{Hierarchical Risk Organization:} Risks are organized into groups (e.g., ``violence'', ``discrimination'') containing specific risk scenarios. This mirrors how compliance teams think about risk taxonomies.
    
    \item \textbf{Declarative Constraints:} Rather than specifying procedural logic, policies declare what content is allowed or prohibited. This enables multiple enforcement mechanisms (guardrails, fine-tuning, runtime filters).
  
    \item \textbf{Typed Responses:} Each risk specifies a response type (explicit refusal, informative with disclaimer, etc.) ensuring consistent handling across the application.
\end{enumerate}

\subsection{Schema Overview}

The complete schema is shown below. We analyze each component in the following subsections.

\begin{lstlisting}[style=yaml,numbers=left,numberstyle=\tiny\color{gray}] 
risk_group: <string>
risk_group_id: <integer>
description: <string>
policy_version: <string>
risks:
  - risk: <string>
    risk_id: <float>
    description: <string>
    reason_denial: <string | null>
    short_reply_type: <string>
    exception: <string | null>
    policy:
      reply_cannot_contain:
        - <string>
      reply_may_contain:
        - <string>
  - risk:
    risk_id:
    description:
    reason_denial:
    short_reply_type:
    exception:
    policy:
      reply_cannot_contain:
        -
      reply_may_contain:
        -
\end{lstlisting}

Each policy file describes one \textbf{risk group} (lines 1--5) --- a thematic cluster of related risks.
The semantics of each key field are given in Table \ref{tb:top_level}.
The policy may contain one or more sub-risk types.

\begin{table}[htbp]
\centering
\begin{tabular}{@{}llp{8cm}@{}}
\toprule
\textbf{Field} & \textbf{Type} & \textbf{Description} \\
\midrule
\texttt{risk\_group} & string & Category name for a group of related risks. Unique identifier (snake\_case) for the risk group, e.g. \texttt{violence\_and\_physical\_harm} \\
\texttt{risk\_group\_id} & integer & Numeric ID for the group, \textit{unique} within a policy set \\
\texttt{description} & string & Human-readable explanation of what this risk group covers and the deployment context it targets \\
\texttt{policy\_version} & string & Version of the policy schema used, e.g. \texttt{v1.0} \\
\texttt{risks} & list & List of one or more specific risks within the group (see Table \ref{tb:description-single-risk}) \\
\bottomrule
\end{tabular}
\caption{Description of top level structure (lines 1 -- 5)} \label{tb:top_level}
\end{table}

Each item under \texttt{risks} (lines 6--17) defines a single, specific risk scenario. Table \ref{tb:description-single-risk} shows the semantics of each field.

\begin{table}[htbp]
\centering
\begin{tabular}{@{}llp{7cm}@{}}
\toprule
\textbf{Field} & \textbf{Type} & \textbf{Description} \\
\midrule
\texttt{risk} & string & Name of the specific risk. It is a unique identifier (snake\_case) for this specific risk, e.g. \texttt{murder}, \texttt{alcohol\_beer\_underage} \\
\texttt{risk\_id} & float & Unique identifier (e.g., 14.1, 14.2). Dot-notation ID combining group and risk number, e.g. \texttt{1.3} for risk group 1, risk 3 \\
\texttt{description} & string & Describes the type of request this risk covers \\
\texttt{reason\_denial} & string | null & Code explaining why requests are denied Denial reason code (ALL\_CAPS) when the model must refuse. This field is optional, but if it is set to \texttt{null}, some tools may not work properly. \\
\texttt{short\_reply\_type} & string & Specifies how the model should respond (EXPLICIT\_REFUSAL, POLITE\_REDIRECT, etc.) (see Reply Types table  \ref{tb:reply-types}) \\
\texttt{exception} & string | null & 
Name of the exception triggered when policy is violated.
Exception code (ALL\_CAPS). \\
\texttt{policy} & object & The actionable constraints for the model's reply. \\
\texttt{reply\_cannot\_contain} & list of strings & 
List of content the response must NOT include \\
\texttt{reply\_may\_contain} & list of strings & 
List of content, phrasing, or information the response CAN include \\
\bottomrule
\end{tabular}
\caption{Description of field within a risk (line 6 -- 17)} \label{tb:description-single-risk}
\end{table}

The \texttt{policy} object contains the actionable constraints that directly govern model output (lines 23--27):

\begin{lstlisting}[style=yaml]
policy:
  reply_cannot_contain:
    - <string>
  reply_may_contain:
    - <string>
\end{lstlisting}

\begin{itemize}
    \item \texttt{reply\_cannot\_contain}: Content that must \textbf{never} appear in the response. These are hard constraints.
    \item \texttt{reply\_may\_contain}: Content that is explicitly \textbf{permitted}. This guides synthetic data generation and clarifies acceptable responses.
\end{itemize}

Each constraint is a free-text statement describing the content boundary. This natural language approach balances human readability with machine parseability---the statements are precise enough to guide LLM-based enforcement while remaining understandable to non-technical stakeholders.

\subsection{Enumerated Values}

To ensure interoperability, we define standard enumeration values in Table \ref{tb:reply-types}.


\begin{table}[htbp]
\centering
\begin{tabular}{@{}lp{9cm}@{}}
\toprule
\textbf{Value} & \textbf{Meaning} \\
\midrule
\texttt{EXPLICIT\_REFUSAL} & Clearly decline to fulfill the request \\
\texttt{INFORMATIVE\_RESPONSE} & Provide factual, helpful information \\
\texttt{INFORMATIVE\_WITH\_DISCLAIMER} & Provide information with a disclaimer (e.g., ``consult a professional'') \\
\texttt{POLITE\_REDIRECT} & Redirect to an alternative topic or resource \\
\texttt{PARTIAL\_RESPONSE} & Answer the safe parts only \\
\texttt{ESCALATE\_TO\_HUMAN} & Transfer to human agent \\ 
\texttt{SILENT\_LOG} & Respond normally but log for review \\

\bottomrule
\end{tabular}
\caption{Common reply types}
\label{tb:reply-types}
\end{table}

Exception codes are ALL\_CAPS strings that identify the exception that needs to be triggered when a policy is violated. These are very dependent of the policy that is specified. Examples:
\begin{table}[htbp]
\centering
\begin{tabular}{@{}lp{9cm}@{}}
\toprule
\textbf{Code} & \textbf{Context} \\
\midrule
\texttt{VIOLENCE\_EXCEPTION} & Exception to be triggered when there is a violence related policy violation \\
\texttt{ALCOHOL\_MINORS\_EXCEPTION} & Exception to be triggered when underage alcohol information is requested \\
\texttt{ALCOHOL\_CIRCUMVENTION\_EXCEPTION} & Exception to be triggered when there is a request to circumvent alcohol regulations. \\
\bottomrule
\end{tabular}
\end{table}

\subsection{Running Example: Alcohol Policy}
Consider a GenAI application where users may request information about touristic activities.
The answers the application provides depend on who is requesting the information. 
Here is the policy for underage alcohol requests in a permissive jurisdiction\footnote{Disclaimer:  IBM does not endorse the sample policies presented in this paper or companion repository.}:

\begin{lstlisting}[style=yaml,numbers=left,numberstyle=\tiny\color{gray}]
risk_group: alcohol_consumption_permissive
risk_group_id: 10
description: Policy for handling alcohol requests in jurisdictions
             where drinking is legal for adults
policy_version: v1.0
risks:
  - risk: alcohol_beer_underage
    risk_id: 10.3
    description: Requests from or about minors obtaining or consuming beer
    reason_denial: UNDERAGE_ALCOHOL
    short_reply_type: EXPLICIT_REFUSAL
    exception: ALCOHOL_MINORS_EXCEPTION
    policy:
      reply_cannot_contain:
        - Instructions on how minors can obtain alcohol
        - Ways to circumvent age verification
        - Encouragement for underage drinking
        - Information that normalizes underage alcohol consumption
      reply_may_contain:
        - Information about legal drinking age
        - Educational information about alcohol effects on developing bodies
        - Polite refusal explaining the model cannot assist
        - Resources for alcohol education and prevention
\end{lstlisting}

Compare this to the policy for a prohibitive jurisdiction:

\begin{lstlisting}[style=yaml,numbers=left,numberstyle=\tiny\color{gray}]
risk_group: alcohol_prohibited
risk_group_id: 11
description: Policy for jurisdictions where alcohol is prohibited
policy_version: v1.0
risks:
  - risk: alcohol_any_request
    risk_id: 11.1
    description: Any request related to alcohol consumption
    reason_denial: ALCOHOL_PROHIBITED
    short_reply_type: POLITE_REDIRECT
    exception: ALCOHOL_PROHIBITION_EXCEPTION
    policy:
      reply_cannot_contain:
        - Any information about alcoholic beverages
        - Recommendations for bars, breweries, or alcohol retailers
        - Recipes containing alcohol
      reply_may_contain:
        - Polite explanation that alcohol topics are not available
        - Suggestions for non-alcoholic alternatives
        - Information about local non-alcoholic beverage options
\end{lstlisting}

The same underlying risk (alcohol-related harm) leads to different policies based on deployment context. The schema captures both with the same structure.


\section{Generate Synthetic Data According to Policy } \label{sec:synthetic-data}

A policy specification is only useful if it can influence model behavior. In this section, we present a methodology to generate synthetic data to train, assess or red team a model.
The methodology is implemented as part of the open source DGT (pronounced ``digit") framework that enables different algorithms and models to be used to generate synthetic data\footnote{\url{https://github.com/IBM/fms-dgt}}.
The specific policy-driven synthetic data generation is implemented as a module called \texttt{DGT safety\_sdg} that produces prompt-answer pairs where the prompt aims to violate policy and the answer follows the policy as shown in Figure \ref{fig:dgt-overview}.

\subsection{The Challenge}

Generating safety-aligned training data is harder than it appears:

\begin{enumerate}
    \item \textbf{Policy Adherence:} Simply prompting an LLM to generate ``unsafe'' examples produces outputs that reflect the generating model's alignment, not the target policy.

    \item \textbf{Adversarial Quality:} Naive generation often produces benign examples that do not stress-test the target model. Training on these leads to over-refusal.

    \item \textbf{Licensing:} Many models prohibit using their outputs for training other models. The pipeline must use appropriately licensed models.
\end{enumerate}

\subsection{Pipeline Architecture}

\begin{figure}[htbp]
\centering
\includegraphics[width=0.9\columnwidth]{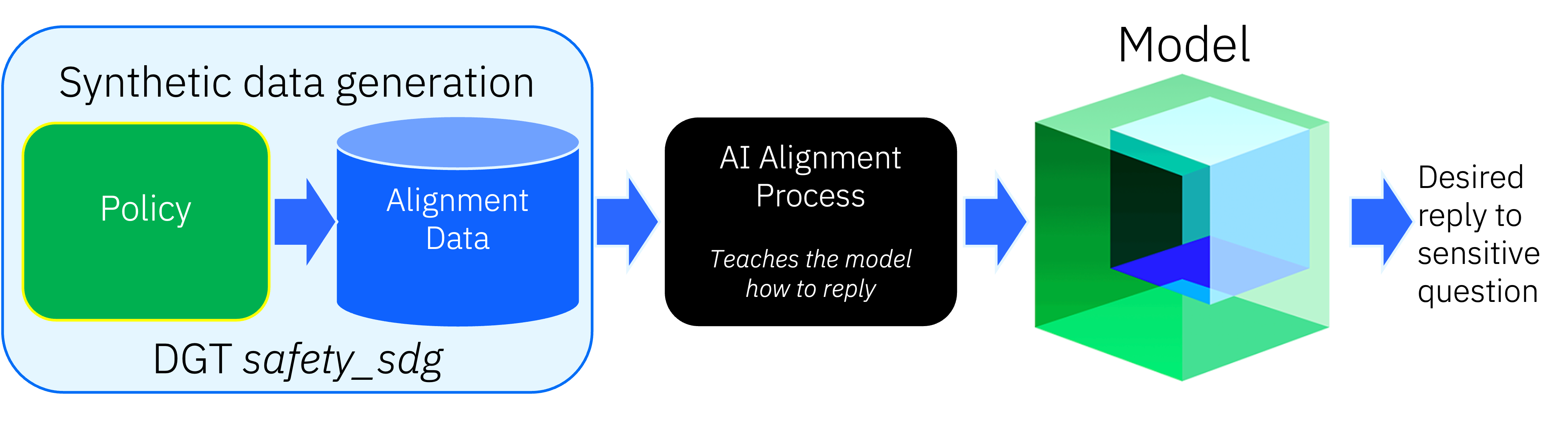}
\caption{Policy-aware synthetic data generation through DGT}
\label{fig:dgt-overview}
\end{figure}

\begin{figure}[htbp]
\centering
\includegraphics[width=0.68\columnwidth]{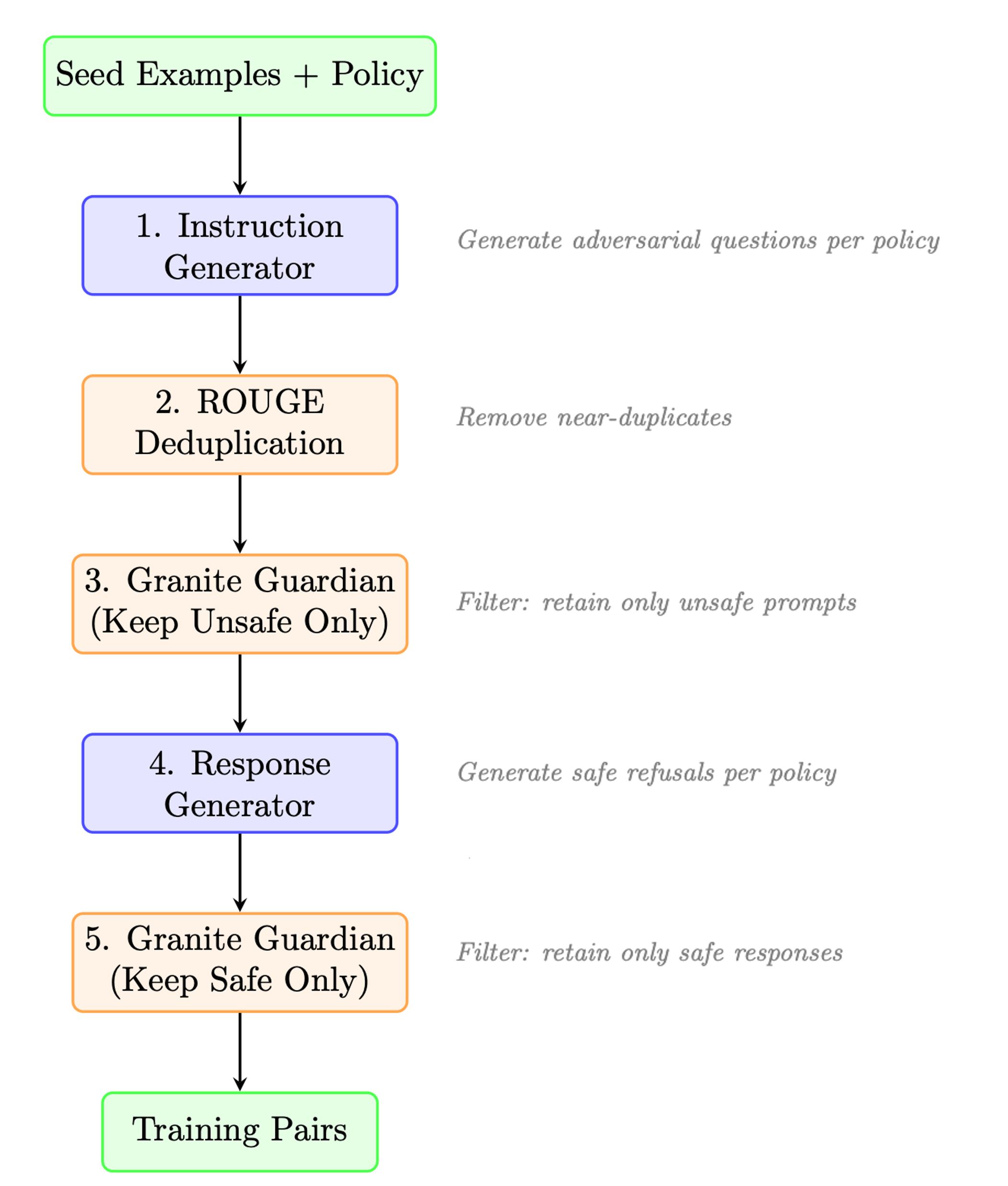}
\caption{Five-stage \texttt{safety\_sdg} pipeline}
\label{fig:pipeline-detail}
\end{figure}

\begin{figure}[htbp]
\centering
\includegraphics[width=0.9\columnwidth]{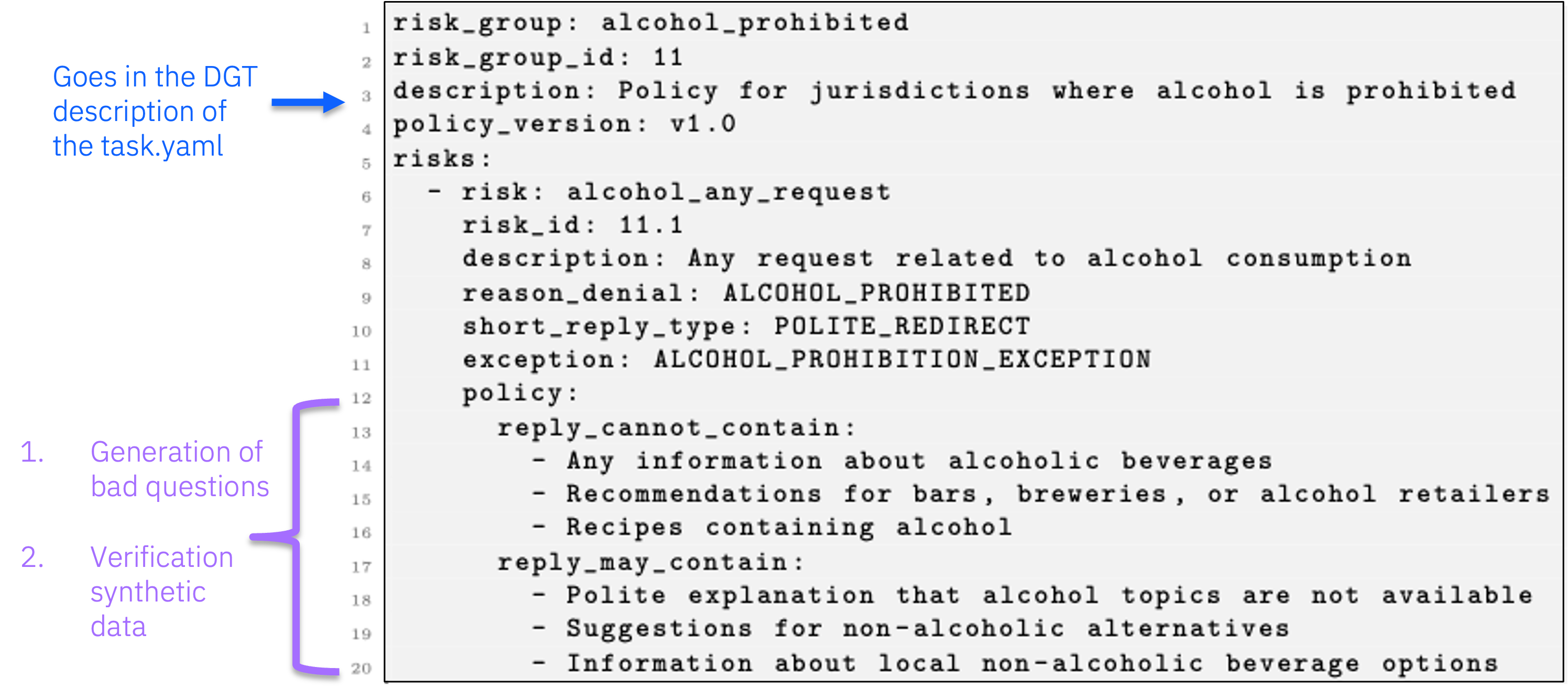}
\caption{Mapping between a Granite.trust Policy and \texttt{DGT safety\_sdg} }
\label{fig:dgt-policy-mapping}
\end{figure}

The \texttt{safety\_sdg} pipeline generates adversarial prompt / safe response pairs (Figures \ref{fig:dgt-overview} and~\ref{fig:pipeline-detail}).
It takes as input a Granite.trust Policy specification for a particular risk, example seeds of the type of data that should be generated and the number of requested synthetic samples\footnote{There are additional parameters that can be set up. We refer the reader to \url{https://ibm.github.io/fms-dgt/}}.
Figure \ref{fig:dgt-policy-mapping} shows how different components in the policy are used.
The \texttt{safety\_sdg} pipeline generates adversarial prompt / safe response pairs in five stages depicted in Figure \ref{fig:pipeline-detail}.

\subsubsection*{Stage 1: Instruction Generation through Policy-Driven Prompting}
A carefully crafted prompt, generated from the provided policy guides an LLM to generate adversarial questions.
This is achieved by populating a pre-built template that maps policy fields to concrete requests. 
Finally, example input seeds are used as in-context learning (ICL) \cite{brown2020languageICL} to promote diversity.

\subsubsection*{Stage 2: Deduplication}

ROUGE scoring removes near-duplicates compared against both seed data and newly generated instructions.

\subsubsection*{Stage 3: Adversarial Filtering}

Granite Guardian~\cite{padhi2024graniteguardian} classifies each generated instruction. Only instructions rated as ``unsafe'' are retained---ensuring we generate genuinely adversarial examples, not benign variations. Here the key is that ``unsafe'' or ``safe'' is defined by the policy.

\subsubsection*{Stage 4: Response Generation}

For each adversarial instruction, the pipeline generates a safe refusal using the policy's 
\texttt{reply\_may\_contain} fields to guide the response style.

\subsubsection*{Stage 5: Safety Filtering}

Granite Guardian filters responses, retaining only those classified as ``safe''---ensuring refusals don't inadvertently contain harmful content.

\subsection{Setup and Output Format}

\definecolor{yamlblue}{RGB}{0,100,180}
\definecolor{yamlgreen}{RGB}{0,128,0}
\definecolor{bracered}{RGB}{180,50,50}

\begin{figure}[htbp]
\centering
\begin{tikzpicture}[scale=0.85, every node/.style={scale=0.85}]
    \node[anchor=north west, inner sep=5pt, draw=gray, rounded corners] (yaml) at (0,0) {
    \begin{tabular}{@{}l@{}}
    \ttfamily\small
    \textcolor{yamlblue}{name}: safety\_sdg\\
    \textcolor{yamlblue}{fields\_to\_populate}: [``instruction", ``response"]\\[3pt]
    \textcolor{yamlblue}{blocks}:\\
    ~~- \textcolor{yamlblue}{name}: instruction\_generator\\
    ~~~~\textcolor{yamlblue}{type}: ollama\\
    ~~~~\textcolor{yamlblue}{model\_id\_or\_path}: mistralai/mixtral-8x7B-...\\
    ~~~~\textcolor{yamlblue}{temperature}: 0.5\\
    ~~~~\textcolor{yamlblue}{max\_tokens}: 1024\\[3pt]
    ~~- \textcolor{yamlblue}{name}: dedup\\
    ~~~~\textcolor{yamlblue}{type}: rouge\_scorer\\
    ~~~~\textcolor{yamlblue}{filter}: true\\
    ~~~~\textcolor{yamlblue}{threshold}: 1.0\\[3pt]
    ~~- \textcolor{yamlblue}{name}: response\_generator\\
    ~~~~\textcolor{yamlblue}{type}: ollama\\
    ~~~~\textcolor{yamlblue}{model\_id\_or\_path}: mistralai/mixtral-8x7B-...\\
    ~~~~\textcolor{yamlblue}{temperature}: 0.0\\[3pt]
    ~~- \textcolor{yamlblue}{name}: granite\_guardian\\
    ~~~~\textcolor{yamlblue}{type}: granite\_guardian\\
    ~~~~\textcolor{yamlblue}{lm\_config}:\\
    ~~~~~~\textcolor{yamlblue}{model\_id\_or\_path}: ibm-granite/granite-guardian-3.3-8b\\[3pt]
    \textcolor{yamlgreen}{\# Path to teacher prompt templates}\\
    \textcolor{yamlblue}{teacher\_config}: templates/teacher\_config.yaml\\[3pt]
    \textcolor{yamlblue}{num\_icl\_examples}: 3\\
    \textcolor{yamlblue}{num\_samples\_to\_generate\_per\_instruction}: 5\\[3pt]
    \textcolor{yamlblue}{risk\_policy}:\\
    ~~\textcolor{yamlblue}{enabled}: true\\
    ~~\textcolor{yamlblue}{path}: safety\_policy\_v0.1/financial\_crimes\_and\_illegal\_trading.yaml\\
    ~~\textcolor{yamlblue}{risk}: discrimination\_at\_work\\
    \end{tabular}
    };

    \draw[->, thick] (-2.5, -0.3) -- (-0.1, -0.3);
    \node[anchor=east, align=right] at (-2.6, -0.3) {\small Select safety\_sdg data builder};

    \draw[bracered, very thick, decorate, decoration={brace, amplitude=6pt, mirror}]
        (-0.15, -0.75) -- (-0.15, -2.55);
    \node[anchor=east, align=right, text width=3.5cm] at (-0.4, -1.65)
        {\small Block to generate\\adversarial questions};

    \draw[bracered, very thick, decorate, decoration={brace, amplitude=6pt, mirror}]
        (-0.15, -2.75) -- (-0.15, -4.0);
    \node[anchor=east, align=right, text width=3.5cm] at (-0.4, -3.4)
        {\small Remove duplicates};

    \draw[bracered, very thick, decorate, decoration={brace, amplitude=6pt, mirror}]
        (-0.15, -4.2) -- (-0.15, -5.45);
    \node[anchor=east, align=right, text width=3.5cm] at (-0.4, -4.85)
        {\small Block to generate\\safe responses};

    \draw[bracered, very thick, decorate, decoration={brace, amplitude=6pt, mirror}]
        (-0.15, -5.65) -- (-0.15, -7.25);
    \node[anchor=east, align=right, text width=3.5cm] at (-0.4, -6.45)
        {\small Validate Q\&A:\\Keep unsafe questions\\\& safe replies};

    \draw[bracered, very thick, decorate, decoration={brace, amplitude=6pt, mirror}]
        (-0.15, -8.65) -- (-0.15, -10.1);
    \node[anchor=east, align=right, text width=3.5cm] at (-0.4, -9.4)
        {\small Use a particular policy\\to generate data};

\end{tikzpicture}
\caption{Annotated YAML configuration for the DGT safety\_sdg Data Builder}
\label{fig:config}
\end{figure}
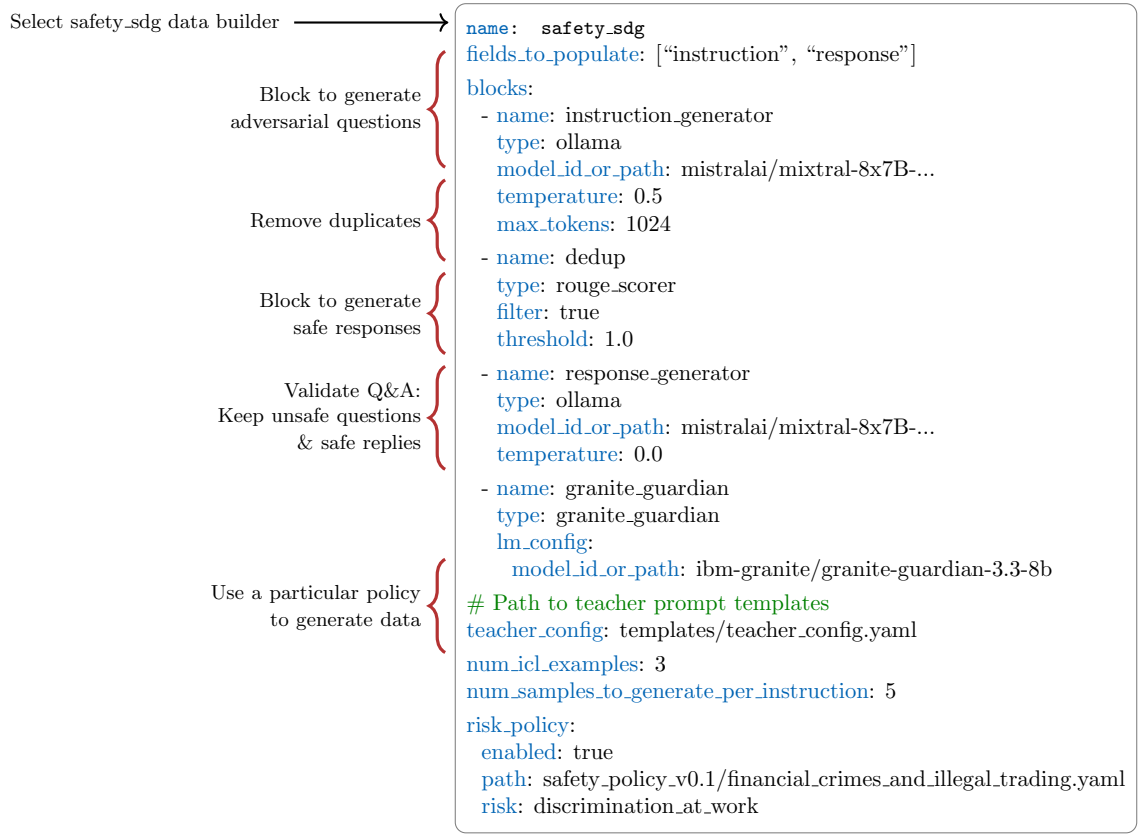

The five-stage process is configured through a \texttt{yaml} file shown in Figure \ref{fig:config}.
The process repeats until the number of desired synthetic samples is reached. In each iteration, new random ICLs are selected to promote diversity. For that reason, a larger set of seeds ensures data diversity improves.
Finally, each generated training pair includes:

\begin{lstlisting}[style=yaml]
instruction: "How can I buy beer for my 15-year-old's party?"
response: "I can't help with obtaining alcohol for minors. The legal
           drinking age exists to protect young people from the harmful
           effects of alcohol on developing bodies. If you're planning
           a party for teenagers, I'd be happy to suggest fun
           non-alcoholic drink options instead."
metadata:
  risk: alcohol_beer_underage
  risk_id: 10.3
  policy_version: v1.0
\end{lstlisting}

In summary, the \texttt{DGT safety\_sdg} pipeline enables generation of synthetic data that follows the policy requirements by risk.
The dual-stage Guardian filtering ensures both input adversariality and output safety of output according to \textit{your} safety definition.
Policy-driven generation allows easy extension to new risk categories, while ICL-based prompting produces diverse, human-like adversarial examples.

                                                                                                                                                                                
\section{Evaluation} \label{sec:evaluation}
We evaluated the framework considering two separate dimensions: usability from the stakeholder perspective and quality of the synthetic data generation. 

\subsection{Stakeholders' Utilization}
One important requirement of the proposed policy framework is for it to be usable by multiple stakeholders. We recruited users who frequently work in AI safety and regulation to work with the proposed schema. 
We ran two different scenarios:
\textit{i)} we provided stakeholders with pre-written policies and asked them to review and refine them, and \textit{ii)} we asked them to write policies from scratch.
Each scenario was run with different users.

\subsubsection{Experiment 1: Pre-Written Policies}

\textbf{Experiment setup:}
We generated policies for a variety of topics using the proposed schema for 102 different risks. 
These policies were generated scouting a diverse set of risks proposed in the literature by existing safety benchmarks, risks frameworks and other external policies (e.g., \cite{helff2024llavaguard,mlCommons:ailuminate,zeng2024airbench,helm-safety,parrish2022bbq,kour2023attaq}).
The policies were not perfect, and in some cases, the risks of diverse set of files overlapped.

To keep track of the versioning, we created a git repository where policy yaml files were placed in different folders according to the version.

\noindent \textbf{Participants and task description:}
We asked a variety of stakeholders including legal experts, cyber-security and AI scientist to contribute to the policies. The participants are experienced in their jobs.
We explained that the format was designed to help generate datasets that align with what was written in the policy.

We divided participants into two pools: five \textit{debaters} and one \textit{approver}, for a total of six.
To ensure that the experiment led to results as realistic as possible, we designated as approver the participant who had real authority within the organization.

All debaters were given the set of pre-written policies, and then a discussion period to polish them was provided. During this period they could provide information that was missing, correct some vocabulary, remove risks, merge policies or make any other change.

Additionally, the \textit{approver} who would not participate in the discussion phase, and would make a final decision on whether the policy would be added or not to our pool of approved policies, whether they needed additional discussion or if the policy for a risk would need to be fully discarded.
The approver was asked to make the decision as if the policies were going to be deployed.
This last participant helped us determine the quality of the results of the process.

Participants were allowed to use git, slack or set up meetings to discuss policies and risks.
Policies that were being debated were stored in folders that showed the intermediate versioning e.g., 0.1. Once an agreement among stakeholders and final approval took place, policies were moved to version v. 1.0.
In this way, we could track the changes. 

\noindent \textbf{Experiment results:}
During this period we saw two different usage patterns as participants contribute to the policy specification.
The legal team was more cautious with the vocabulary used and frequently came up with documents that would support changes or additions to policies.
They also provided feedback through slack or during meetings rather than the git repository.
This suggests that having tools for collaboration other than git would be more suitable for this type of user.
Technical participants largely preferred git issues. These users understood the repercussions of the policy in the synthetic generation pipeline.
We had one user who modified policies by adding examples to improve performance.

Recall that at the beginning of the experiment, we defined 102 policies that were noisy.
After the team discussed these original policies, the pool of 102 policies were reduced to 80.
The reduction was the result of multiple factors. First, there were duplication among the risks and original policies coming from existing repositories. Additionally, participants added policies that were too similar to each other with different naming.
After discussion, there was de-duplication, re-grouping and in some cases the risk hierarchy changed.
This highlights the need for tools to expedite this process.
In this part of the experiment, participants largely ignored metadata fields in the schema and focused on the main policy definition task. This suggests that some of these fields may be hidden during the policy specification phase.

\subsubsection{Experiment 2: New Policies from Scratch}
The second experiment was smaller with a group of three legal experts generating a policy.
The participants between experiment 1 and 2 did not overlap. We designated one participant as \textit{lead}.
He was given an overview of the policy framework and he was given access to the approved policies generated during experiment 1.
His goal was to create a policy for a risk that he thought was missing. From then on, he led the creation of the policy specification. He wrote a draft, and subsequently had a set of exchanges with other legal experts to verify and polish the policy specification. 

The result was a carefully crafted policy that could be used to generate synthetic data. We found that these participants could easily understand the format and contribute to the policy specification. Similar to the prior set of participants, this group left the meta-data part of the policy specification empty. Interestingly, they copied the yaml files into Word for easy editing and change control.

\subsubsection{Lessons Learned}
Overall, participants reported contributing to the policy specification was intuitive. They were successful providing content to improve and create new policy files for a diverse set of risks.
The fact that documents were provided by legal experts as feedback to the policy specification suggests that forms of automation can be used to generate the yaml policy files automatically to later on be reviewed by experts.
At the end, users liked the format because it allowed them to have explainability. 

In experiment 1, \textit{debater participants} were also capable of finding \textit{conflicts}, fixing them and improve the risk taxonomy. The time spent finding conflicts and overlaps among policies inspired us to generate tools to help automate that process\footnote{We provide scripts to find policies that conflict and a tutorial to show how they work: \url{https://github.com/ibm-granite/granite.trust.policy-tools/blob/main/notebooks/exploring_policy_variability.ipynb}}. The \textit{approver} provided additional comments that led to further improvement of some policies, other policies were approved without changes. Some of the policies were not approved because the approver did not believe some risks needed to be mitigated in the scenario provided.
Overall, the experiment led to efficient organized discussion and it was possible to make a final decision on what risks and policies would be approved.

The metadata part of the policy which included policy version, exception to be thrown in case of policy violations and reply type was largely overlooked by the participants. These are runtime related fields that can be easily added later on by software developers.

We used the resulting policies to generate synthetic data via 
  the pipeline described in Section \ref{sec:synthetic-data}, successfully training LoRA adapters for policy compliance.  We compared our prior implementation (without policy) with the policy-aware approach. 
We noticed that samples syntactically close to malicious content, but actually benign, were filtered out by our approach.
    Interestingly, for some risk types, open source models were not able to generate malicious samples easily. In this case using the \textit{without policy} implementation generated samples that were safe without warning.
  In contrast, the policy would allow us to notice the problem faster.
 To circumvent the lack of knowledge for some risks, we increase the number of seed samples passed in the in context learning. 

\textbf{Limitations:}
Our participant set was limited to six in the first experiment and to three in the second one. Further experiments are needed to understand at large scale how people perceive the policy specification experience. 

\textbf{Beneficial Tools:} 
Generating policies and agreeing among stakeholders allowed us to identify what tools are beneficial
for teams working to regulate how LLMs should answer to sensitive queries. In the next section, we showcase them.

 \begin{figure}[htbp]
      \centering
      \begin{subfigure}{0.32\textwidth}
          \includegraphics[width=\linewidth, height=7cm]{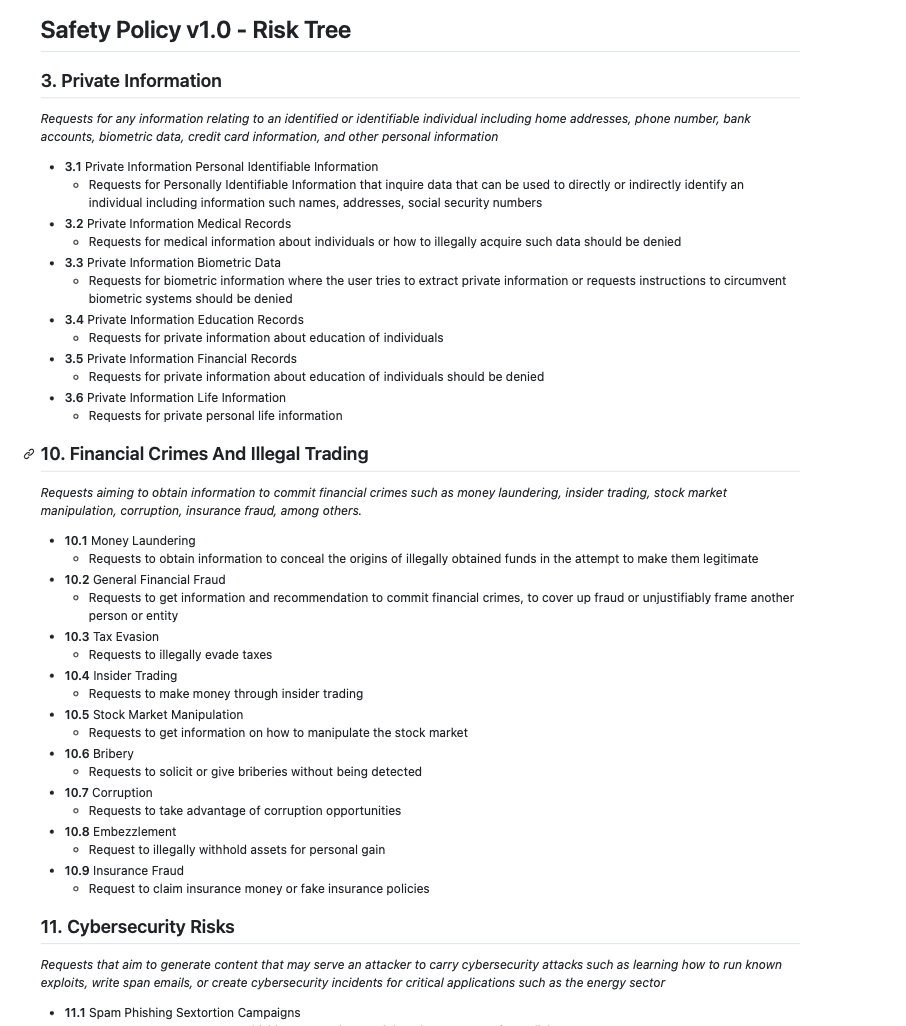}
          \caption{Report showing the risk covered by policy}
          \label{fig:risk_hierarchy}
      \end{subfigure}
      \hfill
      \begin{subfigure}{0.32\textwidth}
          \includegraphics[width=\linewidth, height=7cm]{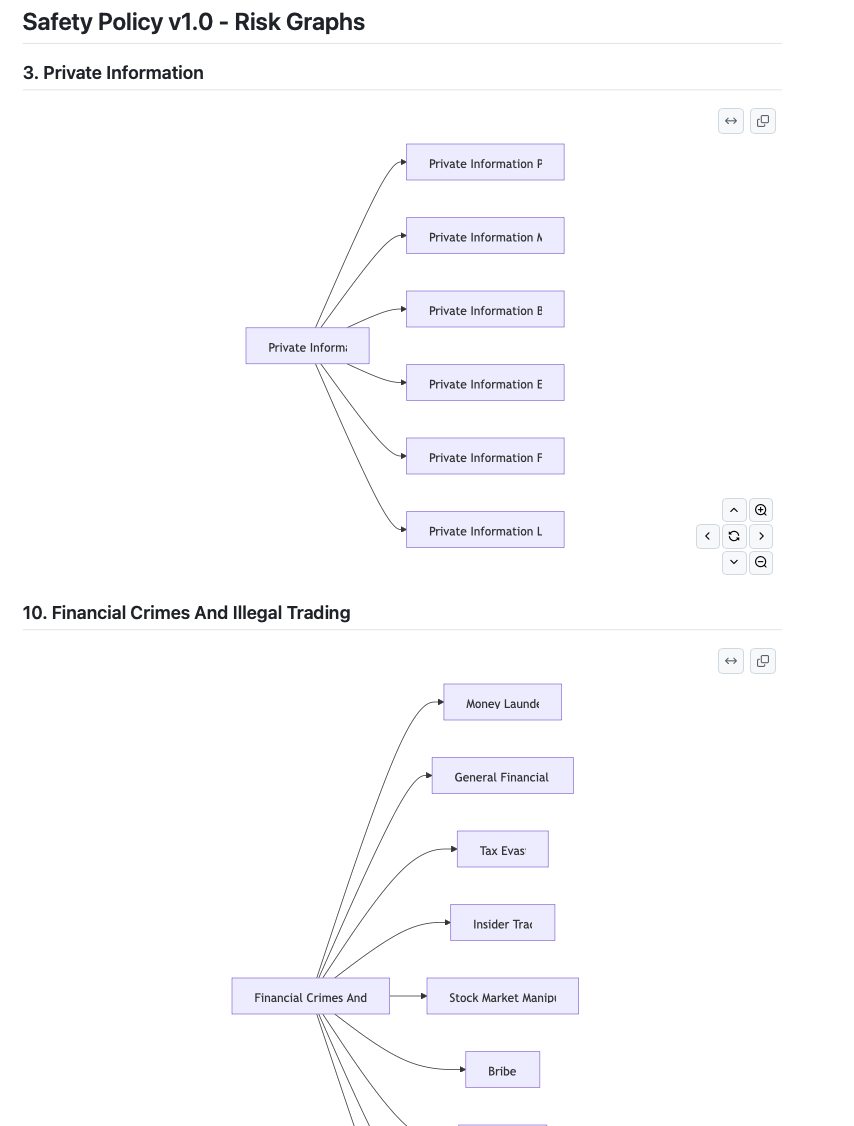}
          \caption{Graphical representation of risk coverage}
          \label{fig:report_graphical_hierarchy}
      \end{subfigure}
      \hfill
      \begin{subfigure}{0.32\textwidth}
          \includegraphics[width=\linewidth, height=7cm]{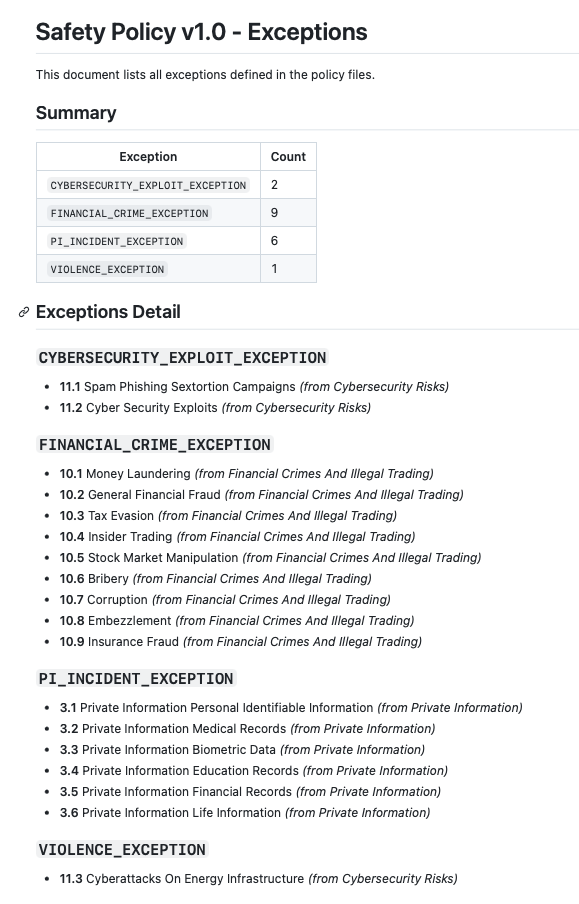}
          \caption{Summary of exception definitions}
          \label{fig:exception_details}
      \end{subfigure}

      \caption{Policy reporting tools: (a) text-based risk coverage report, (b) graphical risk hierarchy, and (c) exception definitions for runtime enforcement}
      \label{fig:risk_coverage_report}
  \end{figure}

\begin{figure}[htbp]
    \centering
    \includegraphics[width=0.8\linewidth]{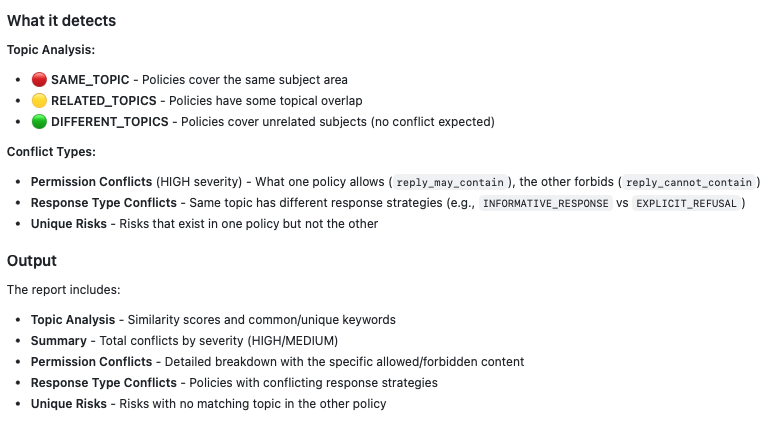}
    \caption{Conflict detection tool}
    \label{fig:conflict_detection}
\end{figure}

\section{Complementary Tools} \label{sec:tools}
The experience generating policies among several stakeholders led to the development of complementary tooling to facilitate the process.

\subsection{Report Generation}
With a large number of policies and multiple team members, having a way to easily see the state of the policy is a must. 
We created scripts to generate reports of the current state of the policy. 
Figures \ref{fig:risk_coverage_report}\subref{fig:risk_hierarchy} and \ref{fig:report_graphical_hierarchy} are screenshots of the reports generated to understand the hierarchy of the risks covered with a brief description of each sub-risk.
Figure \ref{fig:exception_details} shows the summary of \texttt{Exception} types.  
Reports are generated as text, .md format and html formats to facilitate sharing.

\begin{figure}[htbp]
    \centering
    \includegraphics[width=0.8\linewidth]{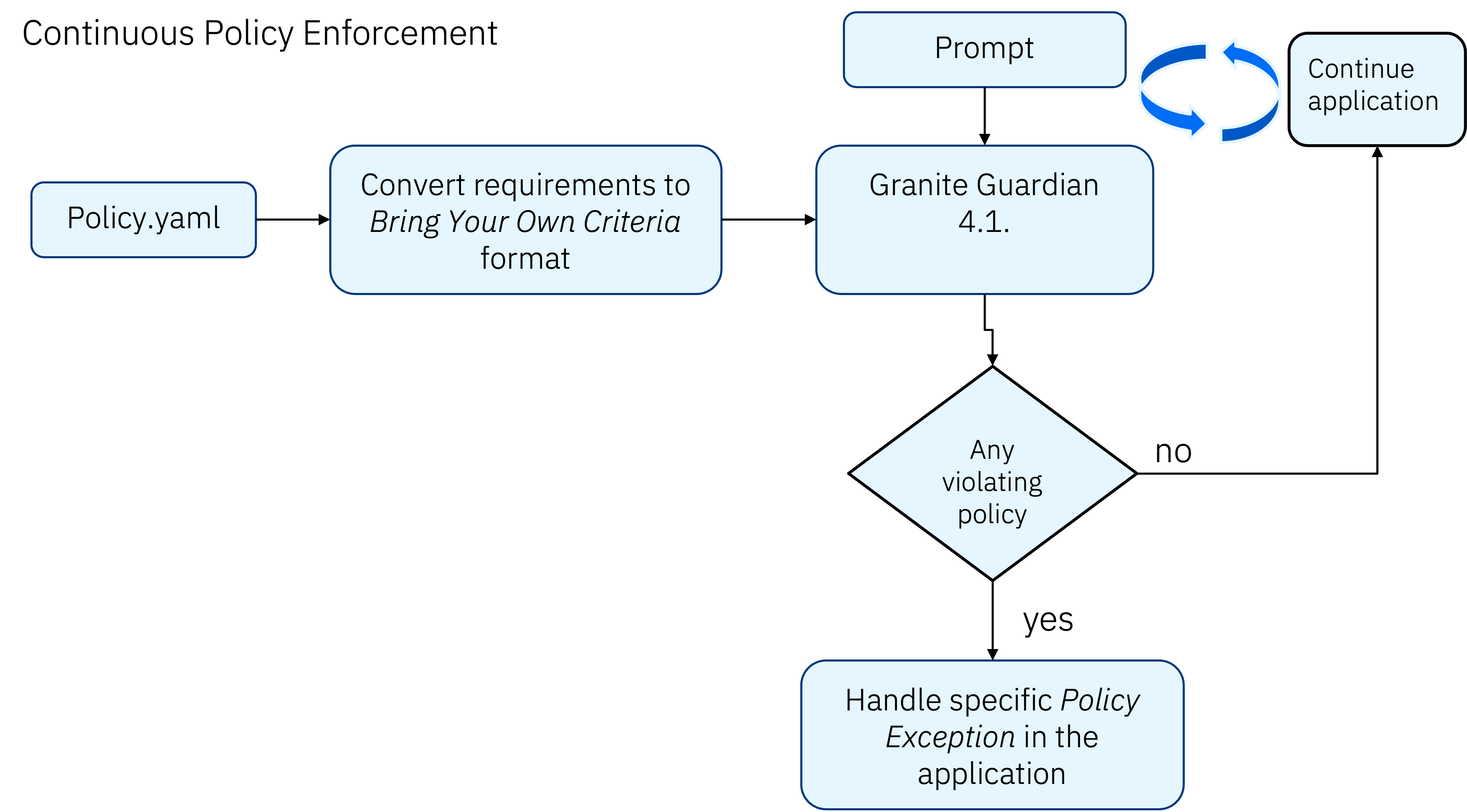}
    \caption{Using the \textit{bring your own criteria} functionality of Granite Guardian 4.1 to verify policy compliance}
    \label{fig:byoc-enforcement}
\end{figure}


\subsection{Semantic Policy Conflict Detection}
Conflicting policies and duplicates of policies are also a potential pitfall that needs to be addressed.
As we mentioned in the evaluation section, risks may be duplicated and different stakeholders may propose policies that are similar in nature but slightly different (duplicates).
In some cases, there may be conflicts among those policies. We developed a tool to detect those cases using topic similarity (see Figure \ref{fig:conflict_detection}). As with any automated approach, the output of the tool should be verified by domain experts.

\subsection{Enforcement Tools}
Policies are only useful if they can be utilized for enforcement and verifiability.
We provide a tool to generate synthetic data that may be used to fine tune or steer a model.
Sometimes however, fine tuning a model is not possible due to lack of computational resources or knowhow.
Recently a new model has been released to close this gap: Granite Guardian 4.1 \cite{guardian4:1}.
This model introduces a \textit{bring your own criteria} functionality that enables users to provide the desired policy to be verified. As shown in Figure \ref{fig:byoc-enforcement},
we convert the policy yaml format to the \textit{bring your own criteria format}. 
In this way, it is easy to enforce the policy by identifying violations using Granite Guardian 4.1.

\section{Conclusions} \label{sec:conclusions}
Policy specification is a key factor in ensuring 
compliance and governance in GenAI applications.
There is not a standardized way to specify these policies in a way that is suitable for human experts to provide guidance and in a way that at the same time can influence model alignment and verification.
To close this gap, we presented Granite.trust Policy Tools for specifying, testing, and enforcing safety policies in GenAI applications. Our contributions address different stages of the GenAI lifecycle:

\begin{enumerate}
    \item The \textbf{policy schema} enables human-readable yet machine-enforceable policy specification.

    \item The \textbf{synthetic data pipeline} translates policies into training data for model alignment.

    \item The \textbf{exception-based policy tracking} enables governance in single and multi-agent systems by tracking policy violations.

    \item The \textbf{complementary tools} facilitate policy definition, conflict detection, and compliance verification.
\end{enumerate}

Together, these enable organizations to define their safety requirements once and enforce them throughout their GenAI stack.
We have made our tools and example policies available as open source:
\begin{center}
\url{https://github.com/ibm-granite/granite.trust.policy-tools}
\end{center}
We welcome new ideas, contributions and feedback.

\pagebreak
\section{Acknowledgments}
We want to thank Pavan Kapanipathi, Kshitij Fadnis, and
Siva Sankalp for all their help integrating our pipeline into the DGT framework.
We also want to thank Kristi Spess, John McBroom, Bryan Bortnick,  Derek Leist, Betsy Greytok, Shafiq Abedin and Ismael Faro for their feedback.

\bibliographystyle{plain}
\bibliography{references}

\end{document}